# Hyperspectral Images Classification and Dimensionality Reduction using spectral interaction and SVM classifier

Asma Elmaizi*, Elkebir Sarhrouni, Ahmed Hammouch, Nacir Chafik

Division of Electronic Systems, Sensor and Nanobiotechnology, ENSET
Mohammed V University, 10000, Rabat, Morocco.

Correspondence should be addressed to Asma Elmaizi; asma.elmaizi@gmail.com

## Abstract

Over the past decades, the hyperspectral remote sensing technology development has attracted growing interest among scientists in various domains. The rich and detailed spectral information provided by the hyperspectral sensors has improved the monitoring and detection capabilities of the earth surface substances. However, the high dimensionality of the hyperspectral images (HSI) is one of the main challenges for the analysis of the collected data. The existence of noisy, redundant and irrelevant bands increases the computational complexity, induce the Hughes phenomenon and decrease the target's classification accuracy. Hence, the dimensionality reduction is an essential step to face the dimensionality challenges. In this paper, we propose a novel filter approach based on the maximization of the spectral interaction measure and the support vector machines for dimensionality reduction and classification of the HSI. The proposed Max Relevance Max Synergy (MRMS) algorithm evaluates the relevance of every band through the combination of spectral synergy, redundancy and relevance measures. Our objective is to select the optimal subset of synergistic bands providing accurate classification of the supervised scene materials. Experimental results have been performed using three different hyperspectral datasets: "Indiana Pine", "Pavia University" and "Salinas" provided by the "NASA-AVIRIS" and the "ROSIS" spectrometers. Furthermore, a comparison with the state of the art band selection methods has been carried out in order to demonstrate the robustness and efficiency of the proposed approach.

## Keywords

Hyperspectral images, remote sensing, dimensionality reduction, classification, synergic correlation, spectral interaction information, mutual information.

## 1. Introduction

With the recent development of the remote-sensing technology [1], the hyperspectral sensors become widely used for the earth monitoring with high spectral resolution. The HSI provide valuable scene observation allowing more accurate material analysis and targets classification. The rich and detailed spectral information provided by the hyperspectral sensors are increasingly used in a variety of commercial, industrial, and military applications. Originally developed by the National Aeronautics and Space Administration (NASA) for remote sensing and space applications, hyperspectral imaging technology is currently used on diverse domains such as agriculture, biotechnology, environmental monitoring and medical diagnostic [2].

The work presented in this paper is focused on the hyperspectral remote sensing applications intending to discriminate the earth surface substances and classify ground objects and materials. The hyperspectral sensors scan the target surface and collect hundreds to thousands of narrow images [3] in the visible and non-visible light spectrum with higher and lower frequencies including the X-rays, infrared and radio-wave all along the electromagnetic spectrum in the form of hyperspectral cubes as illustrated in figure 1. Each image or spectral band represents a narrow range of wavelengths in the electromagnetic spectrum. The bands are combined to form a 3D hyperspectral data cube with two spatial dimensions and one spectral dimension. The figure 1 illustrates an example of different composite substances reflectance of the hyperspectral image "Aviris Indian Pines" [4]. Each pixel in the hyperspectral cube has a corresponding spectral signature that characterizes this specific region of the earth. Thus, every material of the earth surface is identified with its unique electromagnetic spectral signature or reflectance.

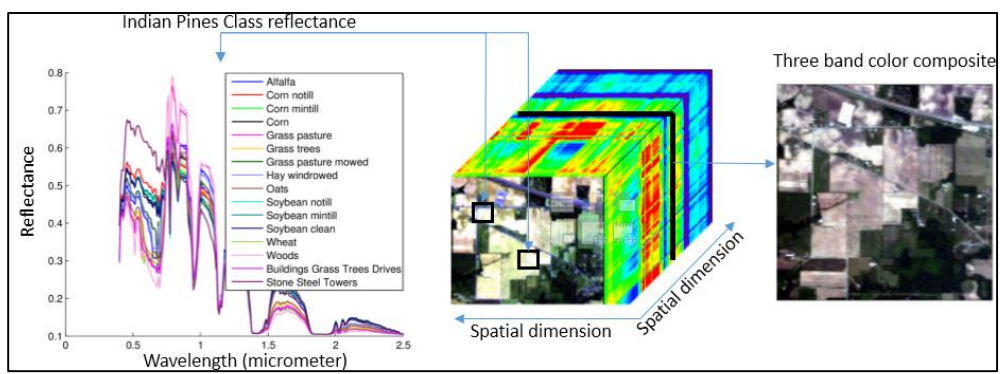

Figure 1. Illustration of the hyperspectral cube of the Indiana Pines and its substances spectral signatures

The remote sensing aims to detect the composition of a specific region using the original spectral signature of the existing materials. The power of this technology lies in the combination of the spatial and spectral characteristics to provide an accurate object classification and detailed information about the supervised area. However, the high dimensionality of the HSI is one of the main challenges for the analysis and processing of the collected data. On the one hand, the high dimensionality brings data storage and computations processing issues [5]. The extraction of useful information in large spaces containing redundant and noisy bands becomes a very challenging task. On the other hand, increasing the number of spectral bands decreases the classification parameters. Previous research had proven that the high dimensionality of HSI leads to classification performance deteriorating, referred in the literature as the Hughes phenomenon or Bellman curse of dimensionality [6][7].

Consequently, several dimensionality reduction approaches have been developed to overcome the computing complexity and the curse of dimensionality challenges. In the literature, the dimensionality reduction techniques have followed two general orientations: feature extraction or transformation and feature selection. The extraction aims to reduce the dimension by projecting the data into a small vector subspace. Several feature extraction approaches have been developed and used in the literature as the example of the principal component analysis (PCA) [8], the local Fisher discriminant analysis (LFDA) [9], the non-parametric weighted feature extraction and discriminant analysis [10]. The main advantage of the extraction approaches is the reduced low-order subset produced after the transformation. However, the transformation of the original datasets could distort the critical information required to discriminate the scene materials and objects in the case of hyperspectral applications.

Besides feature extraction, band selection techniques aim at reducing the original dimension through the selection of a subset of features from the original dataset [11] in order to preserve the original characteristics and features. The selection approaches are divided into three main categories [12]:

- **The Wrapper** is a classifier dependent approach using the classification algorithm as an evaluation function during the selection process. The basic idea is to generate candidate subsets of bands and evaluate them using the classification algorithm as represented by the block diagram in figure 2. The best subset will be hence selected according to the highest classification rate.

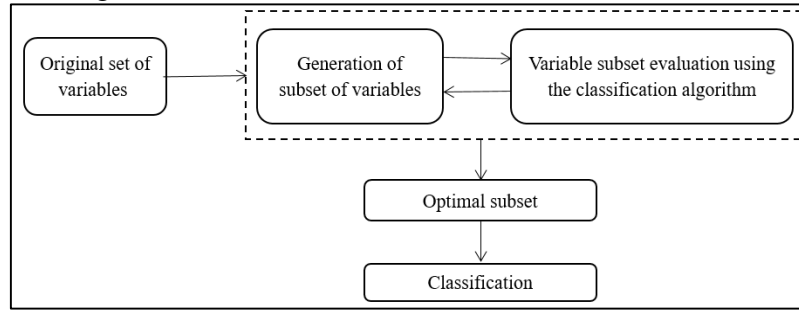

Figure 2. Principle of the Wrapper approach

- **The Filter** approaches are based on the selection of variables that optimizes a defined objective function or model. They are generally independent from the classification algorithm as illustrated in figure 3. The filter approaches classify variables according to their individual predictive power which can be estimated by several measures: information, distance...

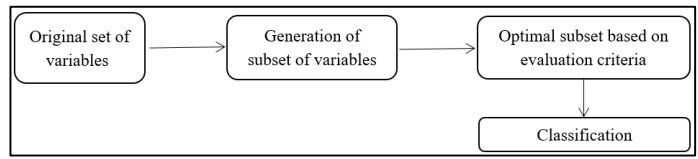

Figure 3. Principle of the Filter approach

- **The Embedded** approaches integrate the variable selection directly in the classification process. Decision trees are the most illustrative example of this approach. The variables are selected at the division level of each node.

**Our proposed development Approach:**

On the one hand, the feature transformation methods reduce the dimension significantly by projecting the data into a small vector subspace. However, the transformation of the original datasets could distort the critical information required to discriminate the scene materials and objects in the case of hyperspectral applications. On the other hand, the selection methods based on wrapper approaches do not provide any theoretical justification for the selection process. The selected subset is specific to the classification algorithm used, it may not be valid if the classification algorithm is changed. Another disadvantage of these methods is the risk of overlearning which usually occurs when the number of instances is insufficient. Additionally, the induction of the classification algorithm increases the computation complexity and influence the selection time.

The filter selection methods are generally less expensive in computation time due to the independency of the selection process to the learning algorithm. In the hyperspectral imaging, filter methods are widely used in band selection for their simplicity, computational efficiency, and independence from the classification strategy. However, the common disadvantage of these methods is the overestimation of the relevance of the bands, which leads to the non-distinction of useful and non-relevant redundancy. Additionally, the lack of information on the interaction between bands influences the selection strategy. The dimensionality reduction using band selection approaches will be the main focus of the work presented in this paper. Our main objective is the development of new approach for the removal of redundant, noisy and irrelevant spectral bands by reducing the original dimensions of hyperspectral datasets while preserving the original cube characteristics and features.

In this paper, we propose a new filter approach for band selection based on interaction information measurement to overcome the limitations of the state-of-the-art filter methods. The work novelty lies on the optimization of the inter-band synergy and interaction information measures to improve the image object classification. The paper is structured as follows: the following part 2 covers the fundamentals of the information theory and literature review of the feature selection state of the art filters based on information measures. The part 3 presents the theoretical foundation and the proposed algorithm (MRMS). The part 4 covers the analysis and discussion of the experimental results carried out on three benchmark hyperspectrale datasets "Indiana", "Pavia" and "Salinas" captured by the NASA AVIRIS [4] and ROSIS [32] and that will be used for the evaluation and assessment of the proposed approach versus the state of art methodologies.

## 2. Hyperspectral bands selection based on information theory

In the literature of feature selection, several approaches based on information theory [13][14] were developed using information measure including: mutual information [15], interaction information [16], conditional mutual information [17] and joint mutual information [18]. Shannon's entropy [13][14] of a variable X is defined as follows:

$$\mathbf{H(X)} = -\sum_{x \in X} \mathbf{p(x) \log(p(x))} \qquad (1)$$

Where P(x) is the probability density function. The joint entropy H(X,Y) between the two variables X and Y is defined as:

$$\mathbf{H(X,Y)} = -\sum_{x \in X} \sum_{y \in Y} \mathbf{p(x,y) \log(p(x,y))} \qquad (2)$$

P(x,y) is the joint probability function of X and Y. P(x) and P(y) are the marginal probabilities. Mutual information (MI) is the amount of information shared by the two variables X and Y as follows (eq 3):

$$\mathbf{MI(X,Y)} = \mathbf{H(X)} + \mathbf{H(Y)} - \mathbf{H(X,Y)} \qquad (3)$$

Information Gain (IG) or Mutual Information Maximization (MIM) [19][20] is a simple approach based on classifying the attributes according to their mutual information with the class. In the hyperspectral imaging, the mutual information is used to measure the relevance of a selected band. X represents the candidate band (B) and Y the ground truth (GT). Simplicity and low computational costs are the main advantages of this method.

However, the dependency between the selected bands is not considered by this approach. As consequence, some selected bands may contain redundant information. To solve this issue, [21][22] proposed the Mutual Information Based Feature Selection approach (MIFS) to select the most relevant attributes using the MI between the candidate feature and the class taking into consideration the dependency between the candidate feature and the previously selected features (eq 4).

$$\mathbf{MIFS = Argmax(MI(X_i, C) - \beta \sum_{X_s \in S} MI(X_i, X_s))} \quad (4)$$

We apply the MIFS approach for the dimensionality reduction of the hyperspectral data. Xi represents the candidate band (Bi), Xs every selected band (Bs) and C the ground truth (GT).

$$\mathbf{MIFS = Argmax(MI(B_i, GT) - \beta \sum_{B_s \in S} MI(B_i, B_s))} \quad (5)$$

According to the MIFS algorithm analysis (eq 5), it is worth noting that when the number of selected attributes increases, the redundancy term increases significantly compared to the relevance term. In this case, some irrelevant attributes can be selected. This problem has been partially solved in the method of minimum redundancy and maximum relevance (mRMR) method proposed by Peng [23], through the division of the redundancy term by the cardinality of the subset. The redundancy term is distributed over the cardinality |S| of the selected subset of attributes, because the redundancy term becomes larger when the selected subset of attributes becomes larger. This modification allows the mRMR to outperform the conventional MIFS and MIFS-U methods.

$$\mathbf{mRMR = Argmax(MI(X_i, C) - \frac{1}{|S|} | \sum_{X_s \in S} MI(X_i, X_s))} \quad (6)$$

We apply the mRMR approach to reduce the dimensionality of the HSI (eq 7). Xi represents the candidate band (Bi), Xs each selected band (Bs), S the cardinality of the subset of selected bands and C the ground truth (GT).

$$\mathbf{mRMR = Argmax(MI(B_i, GT) - \frac{1}{|S|} | \sum_{B_s \in S} MI(B_i, B_s))} \quad (7)$$

Estèvez [24] proposed as well an improved version of the MIFS, MIFS-U and mRMR approaches: the normalized mutual information feature selection (NMIFS). In this method the MI is replaced by the normalized MI (NMI) to improve the redundancy control criterion ().

$$\mathbf{NI(Bi, BS) = \frac{MI(Bi, Bs)}{\min\{H(Bi), H(Bs)\}}} \quad (8)$$

$$\mathbf{NMIFS = Argmax(MI(B_i, GT) - \beta \sum_{B_s \in S} NI(B_i, B_s))} \quad (9)$$

In the hyperspectral imaging, different variants of normalized mutual information measures have been proposed by Nhaila [25], Sarhrouni [26] to improve the band's selection process.

$$\mathbf{NMI(Bi, VT) = \frac{H(B) + H(GT)}{H(B, GT)}} \quad (10)$$

Another common issue to the methods presented previously is the redundancy term calculation on the basis of the MI measure between the candidate feature and the selected features without any consideration of the classes. Yang and Meyer [27] proposed the joint mutual information (JMI) method to solve this problem by including the conditional mutual information to control the redundancy and relevance with consideration of the class information.

$$\text{JMI} = \text{Argmax} \sum_{X_s \in S} I(X_i, X_s; C) \qquad (11)$$

Elmaizi [28] demonstrated that this method performs well in terms of stability and classification accuracy. We apply the JMI method for the reduction of the hyperspectral data. Xi represents the candidate band (Bi), Xs each selected band (Bs) and C the ground truth (GT).

$$\text{JMI} = \text{Argmax} \sum_{B_s \in S} I(B_i, B_s; GT) \qquad (12)$$

Meyer [29] presented the Double Input Symmetric Relevance (DISR) where the mutual information is replaced by the symmetric relevance measure in the objective function (eq 13).

$$\text{DISR} = \text{Argmax}(\sum_{B_s \in S} \frac{I(B_i, B_s; GT)}{H(B_i, B_s; GT)}) \qquad (13)$$

## 3. The proposed approach: Max Relevance Max Synergy filter MRMS

On the one hand, some of the state-of-the-art filter selection approaches ignore the class information during the selection process and focus more on the redundancy measure between the selected attributes. On the other hand, the overestimation of the importance of some pre-selected features can influence significantly the selection strategy. This scenario occurs when the candidate attribute is correlated with a preselected one, but independent from the rest of the subset features. The value of the objective function could increase in this case due to the overestimated redundancy. Consequently, the attribute will not be selected. Actually, the irrelevant attributes removed during the selection process can contain pertinent information about the class when combined with other attributes. Unintended elimination of this type of features during the selection process can result in a loss of useful information and consequently influence classification performance. To overcome these selection challenges, we propose in the following section a new selection filter methodology based on the simultaneous control of relevance, synergy and redundancy measures for the reduction and accurate classification of the hyperspectral images.

### 3.1. Interaction information measure

**Definition:** Interaction information I(X, Y, Z) has been defined by Jakulin [16][30] as the decrease in uncertainty about Z caused by the union of the attributes X and Y in a Cartesian product [31].

The relationship between the information interaction measure and the conditional information is defined as:

$$I(X, Y, Z) = [H(X) + H(Y) - H(X, Y)] - [H(X|Z) + H(Y|Z) - H((X, Y)|Z] \qquad (14)$$

From the eq.14, we deduce the relationship between the joint mutual information and the interaction information measure:

$$I(X, Y, Z) = I((X, Y), Z) - MI(X, Z) - MI(Y, Z) \quad (15)$$

## 3.2. Development of new band selection strategy based on Interaction information measure

Our objective is to develop a band selection strategy based on the simultaneous evaluation of three information measures: relevance, redundancy and synergy between the selected bands. Our proposed selection algorithm is based on the maximization of an objective function composed of two terms:

### a. Interaction information for redundancy and synergy control

The synergy and redundancy measures of the hyperspectral bands are estimated and controlled through the interaction information measure. In the objective function of our algorithm, the interaction information is calculated according to the equation 15 considering X the candidate hyperspectral band (Bi), Y every preselected band (Bs) and Z the ground truth (GT). The interaction information I(Bi,Bs,GT) is employed for the control of synergy and redundancy Inter bands (eq 16). The ground truth estimated (GTest) is calculated through the average of the pre-selected bands.

$$I(Bi, GTest, GT) = MI((Bi, GTest), VT) - MI(Bi, GT) - MI(GTest, GT) \quad (16)$$

In our proposed selection strategy:

- The interaction information measure is positive when the evaluated band provides information about the ground truth when combined with the previously selected bands. This information cannot be provided by each one of them individually (Synergy and complementarity correlation).
- The interaction information measure is negative when the evaluated band contains redundant information already provided by the bands selected previously (redundancy figure 4a).
- The interaction information measure is equal to zero in the case of independency between the evaluated band and the ones previously selected in the context of the class (Independency figure 4b).

**The interaction information graphical illustration**

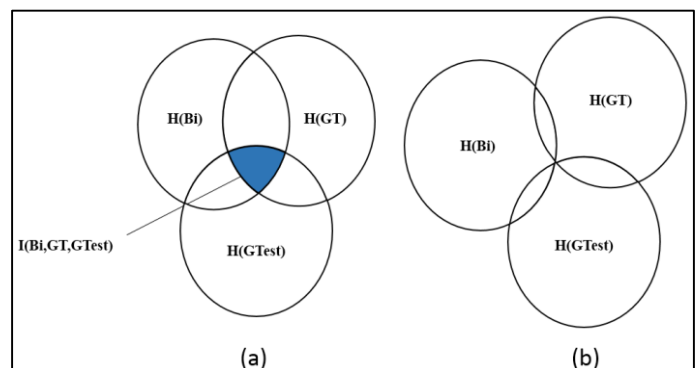

Figure 4. Venn diagrams illustrating interaction information.

### b. Mutual information for relevance control

We introduce the mutual information measure between the candidate band and the ground truth for control of relevance MI(Bi,GT). As illustrated in the block diagram of the proposed approach (figure 5), our objective function Max Relevance Max Synergy (MRMS) (eq17) is defined as a combination of the relevance and the interaction information measures for redundancy and synergy. In every iteration, the algorithm is selecting the band that maximizes the objective function. The maximum number of iterations is equal to the number of the HSI number of bands.

$$\mathbf{MRMS = Argmax(MI(B_i, GT) + I(Bi, GTest, GT))} \quad (17)$$

The MRMS objective function is composed of the terms:

- The measure of band relevance calculated by the MI between the selected band and the ground truth in order to select relevant bands and eliminate irrelevant and noisy bands. This term reflects the information shared between the selected candidate band (Bi) and the ground truth (GT) and will be used to assess the discriminatory capacity of each band in the classification.
- The measure of a band redundancy and synergy correlation will be calculated by the interaction information I(Bi,GTest,GT). This measure controls simultaneously the redundancy and the synergy correlation between the candidate band, the ground truth and the preselected bands for a better classification.

**Block diagram of the proposed MRMS approach:**

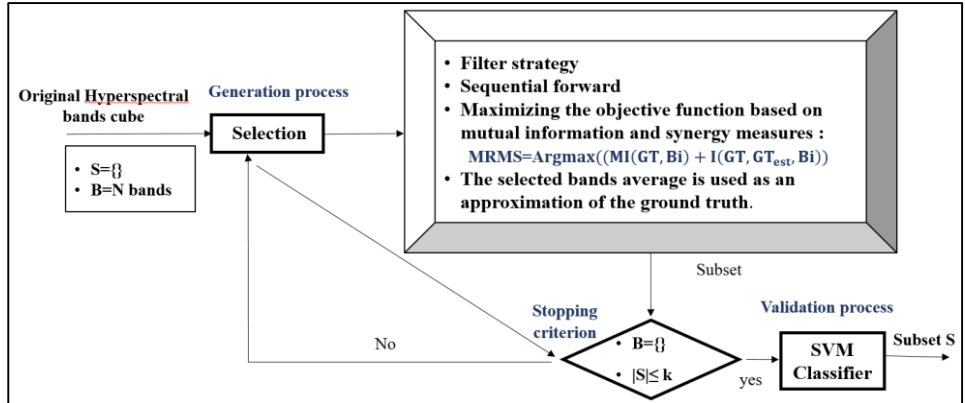

Figure 4. Block Diagram of the proposed approach MRMS

### 3.3. Description of the proposed algorithm MRMS procedure

The MRMS is based on a greedy approach illustrated in figure 5 and 6. The resulting subset of bands S is built up step by step. Before launching the iterations, the first selected band is the one with the largest mutual information measure with the class label or ground truth. During each iteration, the band that maximizes the criterion proposed in (eq.17) is added. The ground truth estimated used in the goal function is recalculated using the average of the subset selected bands at the end of every iteration. After reaching the predetermined number of bands to be selected, the algorithm outputs the optimal subset (S).

| **Proposed Algorithm MRMS** |
|---|

**Input:** B= {B1, B2,…,BN}: Spectral bands of the hyperspectral image, GT: the ground truth , pixels for training , pixels for testing
**Output:** The subset S of k selected bands orderly

1. The initialization of the selection process. Set **S ← [ ]** "empty subset".
2. MI(Bi,GT): calculation of the mutual information measure between each band Bi and the ground truth.
3. Choice of the first selected band: select the band with the largest mutual information measure with the ground truth **B*.**
    a. **B*=Argmax (MI (Bi,G))**
    b. **Set Gest=B***
    c. **Set B←B-{B*], S←{B*}**
4. While |S|≤ k repeat
    $$B*= \mathbf{Argmax}(\mathbf{MI}(\mathbf{Bi}, \mathbf{GT}) + \mathbf{I3}(\mathbf{Bi}, \mathbf{Gest}, \mathbf{GT}))$$
    Choose the band that maximizes the criterion MRMS defined previously in (eq.17)
    Recalculate the ground truth estimated
    a. $\mathbf{Gest} = \frac{\mathbf{Gest} + \mathbf{B*}}{\mathbf{2}}$
    b. **B←B-{B*], S←SU{B*}**
5. After reaching the predetermined number of bands to be selected (k), outputs the subset (S).

Figure 5. Proposed algorithm MRMS selection process

## 4. Experimental datasets, results and analysis

Experimental results have been carried out using diverse hyperspectral datasets captured by the NASA Airborne Visible InfraRed Imaging Spectrometer (AVIRIS) and Reflective Optics System Imaging Spectrometer (ROSIS). The selected images are challenging datasets captured in different types of environments (urban areas, vegetation area), different spatial-spectral resolution and a different number of spectral bands.

### 4.1. AVIRIS dataset: Indiana pines

The Indiana pines is an area of vegetation located north-west of Indiana, USA [4]. The dataset was captured by the Airborne Visible InfraRed Imaging Spectrometer (AVIRIS) sensor. This dataset contains agricultural areas where crops are in the early growth stage with few forested areas and natural perennial vegetation.

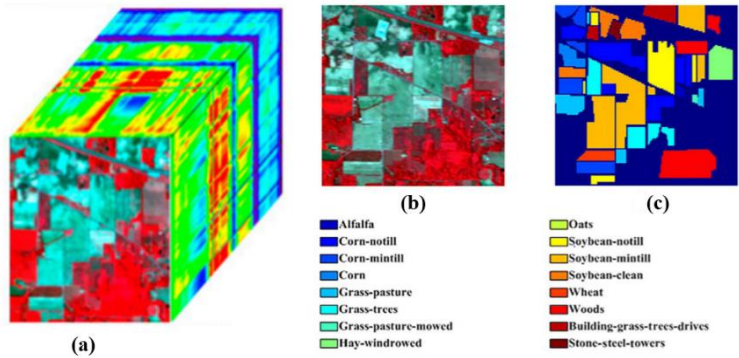

Figure 7. Indian Pines 3D cube (a), three-bands color composite image (b), ground reference (c) .

The Indiana Pines image has a spatial dimension of 145*145 pixels and 20m per pixel spatial resolution. Each pixel is represented in 220 spectral bands covering the range of 375-2500 nm. The ground truth of this image contains 16 labelled classes (figure 7).

## 4.2. ROSIS dataset: Pavia University

The HSI Pavia was captured on the city of Pavia Italy by the Reflective Optics System Imaging Spectrometer (ROSIS) [32]. It represents an urban area of the university of Pavia. The spatial dimensions of the image are 610*340 pixels with 1.3m per pixel special resolution. The hyperspectral cube contains 115 spectral bands with a range of 430 to 860 nm. The dataset consists of different classes: buildings, materials and trees. The ground truth contains 9 classes illustrated in figure 8.

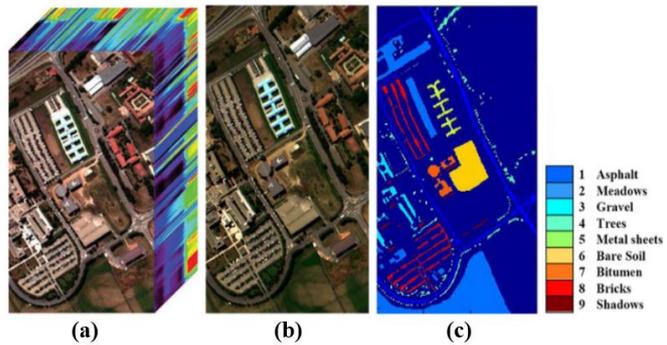

Figure 8. Pavia University 3D cube (a), three-bands color composite image (b), ground reference (c).

## 4.3. AVIRIS dataset: Salinas

The Salinas image was captured on the Salinas Valley located in the USA by the AVIRIS sensor [4]. The image has a spatial dimension of 512*217 pixels with 3.7m per pixel special resolution. Each pixel is represented in 224 spectral bands. The ground truth of this dataset contains 16 classes illustrated in figure 9.

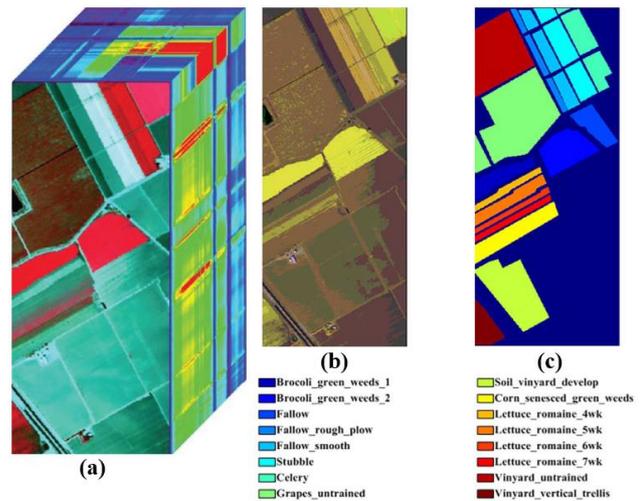

Figure 9. Salinas 3D cube (a), three-bands color composite image (b), ground reference (c).

## 4.4. Parameter setting and classification

To evaluate the performance of the proposed approach, 10%, 25% and 50% of the pixels of each class are randomly selected for training. The remaining samples are used for test and validation. The support vector machine (SVM) classifier [33][34] is used with the radial basis function core (RBF) for the classification. The overall accuracy (OA), average accuracy (AA), kappa (K) coefficient and the specificity (SP) classification metrics will be used to evaluate the performance of the proposed approach.

## 4.5. Experimental Results: Indiana Pines Dataset

In the following part, we present the performance of the proposed approach through the classification metrics: the Individual Class Accuracies (ICA), the Overall Accuracy (OA), the Kappa (K) and specificity (SP) coefficients obtained from the Indiana pines dataset. The table 1 illustrates the overall accuracy result comparison using our proposed approach with different training percentage pixels. 10%, 25%, and 50 % were used for training and the remaining part for testing and validation. According to table 1 results, the size of the training sets increasing leads to the overall accuracy rate improving.

It is remarkable that the classification accuracy rate is higher when we select a subset of bands instead of using all the image bands which confirms the curse of dimensionality phenomena discussed previously [6][7]. The high classification accuracy was obtained using 50% training: 95.87%.

Table 1- Indiana Pines Dataset : the proposed approach (MRMS) overall accuracy results

| Selected bands number | The percentage of pixels for training (%) | | |
|---|---|---|---|
| | 10% | 25% | 50% |
| 60 | 83.97 | 90.00 | 95.07 |
| 70 | 84.65 | 90.60 | 95.49 |
| 80 | 85.06 | 90.85 | 95.87 |

The following table (Table 2) presents the performance comparison results of the proposed approach versus several state-of-the-art filter algorithms: MIBF, MIFS, mRMR, MIH, NMI, JMI and DISR already highlighted in section 2. Table 2 columns refer to the individual class accuracies (ICA) of the captured image. Each row of the table represents 1 of the 16 observed scene classes. The last four rows contain the classification measures calculated for different methods: the kappa coefficient, the specificity, the overall accuracy (OA), the average accuracy (AA).

Table 2- Indiana Pines: results of the proposed approach versus different filters (50 selected bands).

| Class | MIBF | MIFS | mRMR | MIH | NMI | JMI | DISR | Proposed Approach MRMS |
|---|---|---|---|---|---|---|---|---|
| | | | | ICA | | | | |
| 1 | 90.74 | 57.41 | 57.41 | 92.59 | 83.33 | 90.74 | **94.44** | 92.59 |
| 2 | 80.13 | 63.81 | 63.6 | 76.92 | 75.94 | 82.71 | 86.19 | **91.07** |
| 3 | 85.13 | 52.52 | 52.76 | 83.09 | 78.18 | 86.81 | 87.17 | **90.77** |
| 4 | 81.2 | 50 | 50 | 80.34 | 76.5 | 88.03 | 89.74 | **91.88** |
| 5 | 89.94 | 53.72 | 52.31 | 90.54 | 86.32 | 97.79 | **98.39** | 97.79 |
| 6 | 96.12 | 82.33 | 85.01 | 94.51 | 94.51 | 99.2 | **99.2** | 98.8 |
| 7 | 76.92 | 50 | 50 | 69.23 | 73.08 | 88.46 | 88.46 | **92.31** |
| 8 | 97.34 | 73.01 | 73.21 | 97.55 | 96.52 | 98.57 | 98.98 | **98.98** |
| 9 | 90 | 50 | 50 | 75 | 90 | 90 | 90 | **90** |
| 10 | 83.68 | 55.37 | 53.41 | 80.58 | 78.82 | 77.38 | 86.88 | **94.21** |
| 11 | 87.28 | 96.88 | 96.39 | 86.35 | 86.43 | 89.91 | 93.03 | **94** |
| 12 | 90.07 | 51.14 | 50.65 | 89.41 | 85.83 | 83.88 | 92.67 | **93.49** |
| 13 | 97.64 | 60.38 | 56.13 | 98.11 | 97.64 | 99.06 | 99.06 | **99.06** |
| 14 | 95.52 | 94.82 | 94.98 | 95.98 | 94.44 | 98.15 | 98.15 | **98.76** |
| 15 | 65.79 | 57.63 | 57.63 | 62.11 | 63.68 | 85.53 | **87.37** | 85 |
| 16 | 96.84 | 51.58 | 51.58 | 96.84 | 96.84 | 96.84 | **97.89** | 96.84 |
| Specificity | 99.1 | 97.9 | 97.89 | 98.99 | 98.9 | 99.23 | 99.45 | **99.59** |
| Kappa | 86.76 | 71.82 | 71.53 | 85.31 | 83.83 | 88.98 | 92.01 | **94.04** |
| AA | 87.77 | 62.54 | 62.19 | 85.57 | 84.88 | 90.81 | 92.98 | **94.1** |
| OA | 87.58 | 73.58 | 73.31 | 86.22 | 84.84 | 89.67 | 92.51 | **94.41** |

The analysis of the results (Table 2, figure 10) demonstrate that our proposed method provides satisfactory classification results compared to the other state of the art mutual information-based approaches. The MIBF, MIFS and mRMR techniques show the lowest classification accuracy rate due to the redundancy control criterion used during the band selection process. The method NMI gives better classification results compared to the approaches MIBF and MIFS due to the improved objective function including the normalization and allowing more control on bands redundancy versus relevancy. The algorithm MIBF and MIFS represents another inconvenient of using the threshold to control the band's redundancy which could lead to the loss of the selection feature of the algorithm in case of a wrong chosen threshold (the threshold TH=-0.02 is chosen experimentally and used for the redundancy criterion). The MIH also gives promising results compared to the MIBF filter. This result could be explained by the homogeneity measure including in this technique to evaluate the relevance of the bands. The MIH is considered the texture and spatial criteria in order to improve the selection process and classification results. The analysis of the comparison results shows that the classification accuracy rate obtained by the proposed algorithm (MRMS) exceeds the other state of art comparison algorithms with (OA=94,41% & Kappa =94.04%, Specificity=99.59%) for 50 selected bands. The methods JMI and DISR give the closest performance to the proposed approach as their evaluation functions are based on the joint mutual information criterion evaluating the joint correlation between the picked bands and the scene ground truth. The MRMS results analysis shows two main advantages of the proposed selection process: considering the selected band's relevance through the relevant criterion and essentially the evaluation of the band's interaction synergy during the selection process. The classification results for each class of the Indiana pines demonstrate the strength of our algorithm in discriminating the material of each class with higher accuracy.

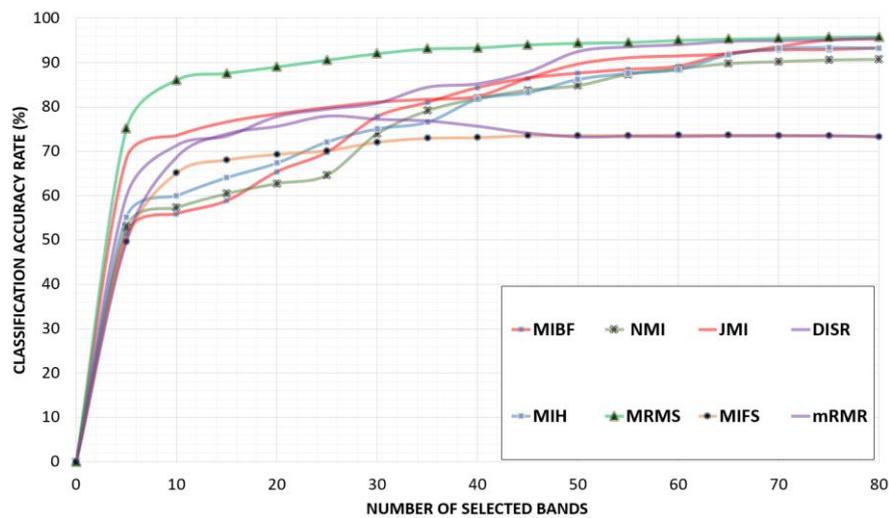

Figure 10. Indiana Pine Dataset : overall classification accuracy comparison results of the proposed approach (MRMS) versus filter methods for a selected number of bands.

Figure 10 illustrates the evaluation and comparison curves of the proposed approach versus the state of art selection algorithms for different selected bands number. The analysis of the obtained results shows the strength of our proposed approach compared with the other mutual information-based approaches. By choosing the discriminative bands rapidly, the proposed approach achieves the highest classification accuracy rate. On the Indiana pines, the MRMS method reaches 94.41% OA overall classification accuracy for 50 selected bands. This result is higher than the closest methods DISR by 2% and more than 4.74 % from JMI approach. Figure 11 illustrates the reproduced ground truth maps using the selected bands obtained by applying our proposed algorithm versus the literature selection filters. We observe that the proposed algorithm attempts to select the smallest set of discriminative bands to classify the image pixels and provide the best approximation of the ground truth and better materials detection.

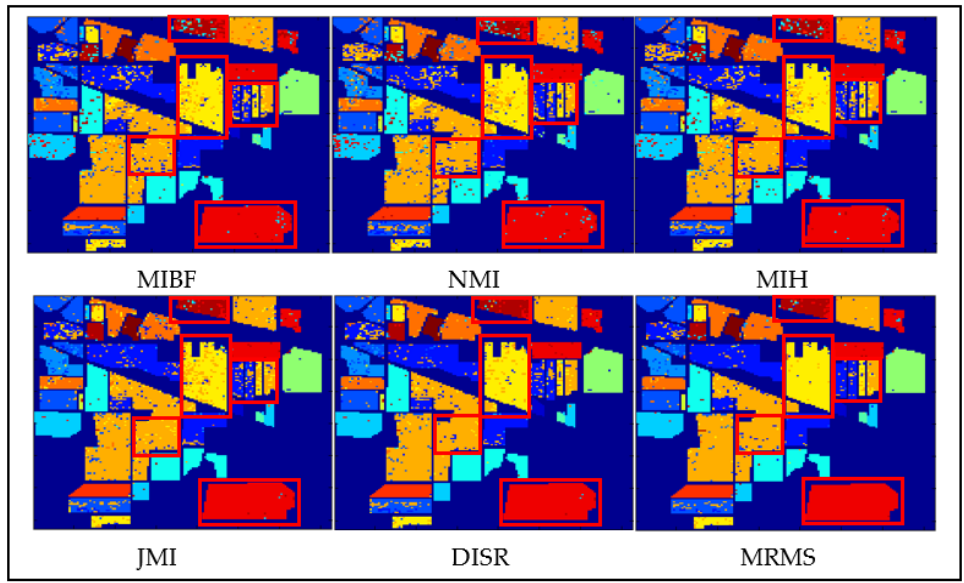

Figure 11. The reproduced ground truth given by the Classification results (Indian Pines image) of the proposed approach (MRMS) versus the MI based filters.

**4.6. Experimental results: Pavia University Dataset**

The following table 3 presents a comparison study between the proposed approach and the mutual information based filter on the Pavia university dataset. Table 3 columns represent the individual class accuracies of the observed scene. Each row of the table represents 1 of the 9 classes. The analysis of the obtained results shows that the proposed method selects bands with high discriminative power very rapidly. The proposed approach achieves an overall accuracy of 94.6% for the first 50 picked bands which is higher than the second-best approaches (DISR & NMI) by 2.2%. The classification accuracy details show that the MRMS algorithm outperforms the MI-based filters in detecting the material and objects of the image.

The figure 12 illustrates the evaluation and comparison curves of the proposed approach and literature MI-based algorithms for different selected band's number on the observed Pavia scene. The results analysis demonstrates that MRMS proposed approach selects pertinent and discriminative bands rapidly due to the strong selection synergy criterion. According to the results, dimensionality reduction by selecting the first 50 bands is sufficient to detect and discriminate the targets and materials in all classes.

Table 3- Dataset Pavia University: results of the proposed approach versus different filters (50 selected bands).

| Class | MIBF | MIFS | mRMR | MIH | NMI | JMI | DISR | Proposed Approach MRMS |
|---|---|---|---|---|---|---|---|---|
| | | | | ICA | | | | |
| 1 | 91.86 | 93.59 | 93.59 | 92.76 | 94.16 | 95.07 | **95.61** | 94.99 |
| 2 | 97.69 | 97.13 | 97.13 | 96.57 | 97.07 | 96.95 | 96.96 | **98.32** |
| 3 | 67.69 | 76.32 | 76.32 | 69.41 | 77.08 | 72.65 | 74.99 | **81.75** |
| 4 | 95.59 | 90.96 | 90.96 | 82.02 | 94.61 | 91.38 | 91.61 | **95.79** |
| 5 | 100 | 99.93 | 99.93 | 99.93 | 100 | 99.93 | 99.93 | **100** |
| 6 | 83.4 | 64.21 | 64.21 | 70.01 | 77.67 | 78.78 | 79.14 | **86.74** |
| 7 | 74.66 | 85.64 | 85.64 | 85.79 | 86.09 | 85.34 | **86.62** | 86.02 |
| 8 | 86.12 | 91.53 | 91.53 | 88.62 | 91.09 | 90.79 | 91.06 | **91.91** |
| 9 | 100 | 99.68 | 99.68 | 99.68 | 99.89 | 99.68 | 99.68 | **99.89** |
| Specificity | 98.84 | 98.48 | 98.48 | 98.39 | 98.85 | 98.81 | 98.86 | **99.2** |
| Kappa | 90.9 | 89.37 | 89.37 | 88.34 | 91.54 | 91.22 | 91.58 | **93.93** |
| AA | 88.59 | 88.78 | 88.78 | 87.2 | 90.85 | 90.06 | 90.62 | **92.82** |
| OA | 91.91 | 90.55 | 90.55 | 89.64 | 92.48 | 92.19 | 92.52 | **94.6** |

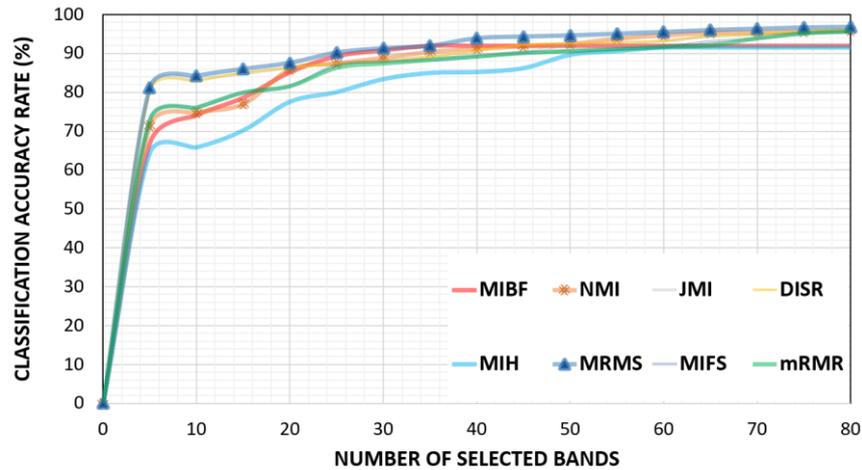

Figure 12. Pavia University Dataset: overall classification accuracy comparison results of the proposed approach (MRMS) versus filter methods for a selected number of bands.

The figure 13 illustrates the reproduced ground maps obtained by the proposed MRMS algorithm versus the MI-based filters. The analysis of the class shows that the MRMS outperforms the MI filters in detecting the objects of the classes and classifying the image pixels with higher accuracy.

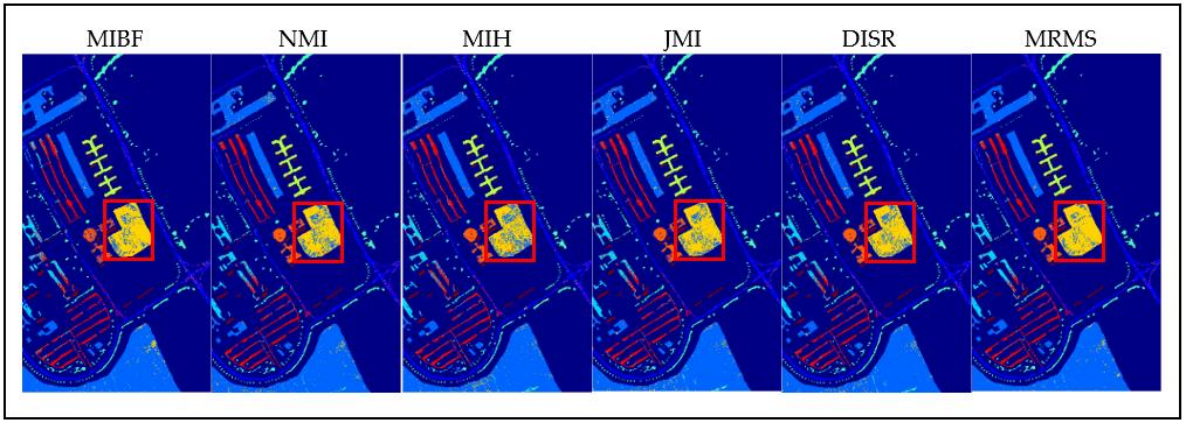

Figure 13. The reproduced ground truth given by the Classification results (Pavia University) of the proposed approach (MRMS) versus the MI-based filters.

### 4.7. Experimental Results Obtained from Salina

Figure 14 illustrates the comparison of the curves results of the proposed approach versus the state of art MI-based filters for different selected band's number on the observed Salinas scene. The curves comparison analysis (figure 14 and table 4) demonstrates the strength of the MRMS proposed approach in selecting the discriminative bands rapidly due to the strong selection synergy criterion.

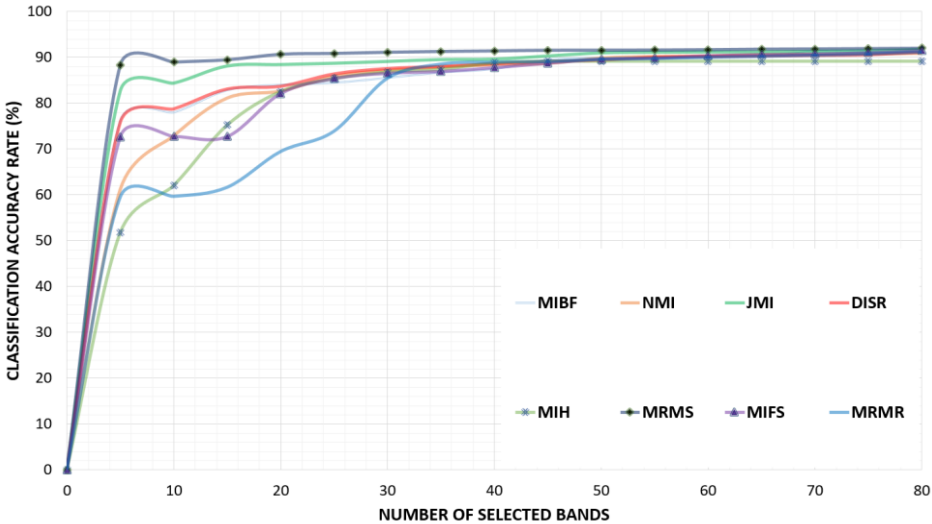

Figure 14. Dataset Salinas: result of the overall classification accuracy of the proposed approach (MRMS) versus the number of selected bands.

By choosing the discriminative bands rapidly, the MRMS filter achieves the highest classification accuracy. On Salinas dataset (table 4), MRMS achieves 91.41% classification accuracy for 40 bands, that is higher than the closest method JMI by 1.72% % and more than 3% from the DISR filter. The Salinas results are aligned with the previous analysis obtained from Indiana and Pavia.

Table 4- Dataset Salinas: results of the proposed approach versus different filters (40 selected bands).

| Class | MIBF | MIFS | mRMR | MIH | NMI | JMI | DISR | Proposed approach MRMS |
|---|---|---|---|---|---|---|---|---|
| | | | | ICA | | | | |
| 1 | 94.47 | **99.75** | 99.35 | 96.76 | 96.62 | 97.46 | 98.56 | 99.5 |
| 2 | 92.3 | 99.95 | 100 | 99.73 | 99.89 | 97.56 | 99.97 | **100** |
| 3 | 99.7 | 98.38 | 98.23 | 99.19 | 97.27 | 98.73 | 98.48 | **98.79** |
| 4 | 96.34 | 99.43 | 99.43 | 99.64 | **99.64** | 99.5 | 99.5 | 99.35 |
| 5 | 98.39 | 97.12 | 97.76 | 98.62 | 97.61 | 96.71 | 96.79 | **98.47** |
| 6 | 99.07 | 99.87 | 99.87 | 99.97 | 99.97 | 99.12 | 99.97 | **99.95** |
| 7 | 88.13 | 99.50 | 99.58 | 99.36 | 99.72 | 99.75 | **99.75** | 99.72 |
| 8 | 88.87 | 90.20 | 88.78 | 88.49 | 89.1 | 89.73 | **91.89** | 89.73 |
| 9 | 99.27 | 99.19 | 99.85 | 99.45 | 99.27 | 99.44 | 99.24 | **99.87** |
| 10 | 91.21 | 86.82 | 86.09 | 91.61 | 90.82 | 89.99 | 91.58 | **94.11** |
| 11 | 95.04 | 87.27 | 87.45 | 90.17 | 90.17 | 89.42 | 91.57 | **99.06** |
| 12 | 98.91 | 99.79 | 99.58 | 98.81 | 98.6 | 98.34 | 99.07 | **99.79** |
| 13 | 98.47 | 99.02 | 99.34 | 98.58 | 99.02 | 98.58 | **99.13** | 98.03 |
| 14 | 91.4 | 93.93 | 94.95 | 93.64 | 93.93 | **95.14** | 93.46 | 93.93 |
| 15 | 45.2 | 36.17 | 46.82 | 45.73 | 41.99 | 51.51 | 37.42 | **57.79** |
| 16 | 94.85 | 98.62 | 98.67 | 92.53 | 94.58 | **98.78** | 96.79 | 98.17 |
| Specificity | 99.06 | 99.07 | 99.16 | 99.15 | 99.12 | 99.22 | 99.13 | **99.35** |
| Kappa | 86.59 | 86.97 | 88.26 | 88.08 | 87.59 | 89 | 87.8 | **90.83** |
| AA | 91.98 | 92.81 | 93.49 | 93.27 | 93.01 | 93.74 | 93.32 | **95.39** |
| OA | 87.43 | 87.78 | 88.99 | 88.82 | 88.36 | 89.69 | 88.56 | **91.41** |

Figure 15 summarizes the results presented previously for the three studied datasets and demonstrate the robustness of the proposed approach. The literature MI-based algorithms eliminate non-pertinent bands based on their mutual information with the ground truth of the observed scene and neglect the complementarity and synergy between the picked bands. The good performance of the proposed approach MRMS compared to the state of art MI-based algorithms demonstrate the importance of the synergy and interaction between the selected subset bands in order to the relevance, redundancy and provide an accurate classification.

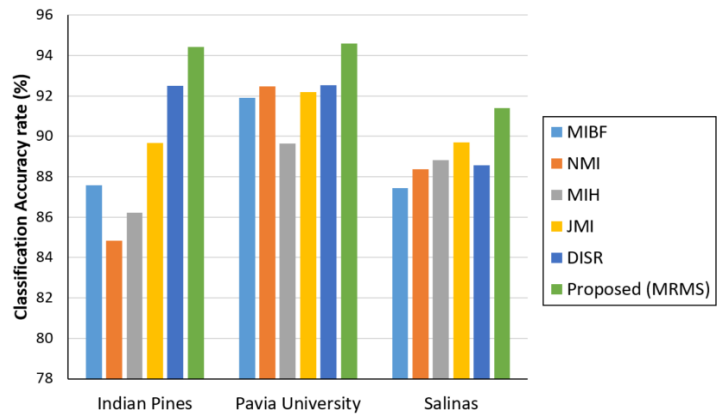

Figure 15. Classification average accuracy bars summary results of Indian Pines, Pavia University and Salinas datasets.

## 5. Conclusion

In this paper, we propose a novel hyperspectral images selection approach based on relevance and synergy maximization (MRMS) to deal with the challenge of hyperspectral images dimensionality reduction and classification. The main idea is to model an objective function that controls the relevance and spectral interaction and synergy between the selected bands and deal with the band selection literature limitations and challenges. We started with a review of the most popular state-of-the-art filter algorithms that will be used to evaluate and compare the performance of the proposed approach. The experiment was conducted on three benchmarks hyperspectral datasets used in the remote sensing domain and provided by the NASA AVIRIS and ROSIS spectrometers: the Indiana Pines, Salinas and Pavia University. The analysis of the image classification obtained results demonstrates the effectiveness and robustness of the proposed approach that achieves higher classification accuracy compared to the reproduced state-of-the-art filter methods using different datasets. The interaction and synergy between the hyperspectral selected bands is an important criterion that should be taken into consideration in order to select the most discriminative bands rapidly and increase the classification of the image pixels. Thus, reproducing an accurate ground truth estimation. As perspectives, we aim to integrate the spatial measure to the proposed algorithm in order to integrate the spatial characteristics of the bands in the selection process.

## Conflicts of Interest

The authors declare that they have no conflicts of interest.